\documentclass[runningheads]{llncs}

\usepackage{eccv}

\usepackage{eccvabbrv}

\usepackage{graphicx}
\usepackage{booktabs}

\usepackage[accsupp]{axessibility}  

\usepackage{multirow}

\usepackage{hyperref}

\usepackage{orcidlink}

\begin{document}

\title{Scene Depth Estimation from Traditional Oriental Landscape Paintings} 

\titlerunning{ }

\author{Sungho Kang\inst{1}\orcidlink{0009-0008-5829-4503} \and
YeongHyeon Park\inst{1,2}\orcidlink{0000-0002-5686-5543} \and
Hyunkyu Park\inst{1}\orcidlink{0009-0003-4889-1498} \and
Juneho Yi\inst{1}\orcidlink{0000-0002-9181-4784}}

\authorrunning{Kang \textit{et al}.}

\institute{Department of Electrical and Computer Engineering, Sungkyunkwan University, Suwon 16419, Republic of Korea \and
SK Planet Co., Ltd., Seongnam 13487, Republic of Korea\\
\email{sungho369@skku.edu, yeonghyeon@sk.com, mjss016@skku.edu, jhyi@skku.edu}\\}
\maketitle

\vspace{-10px}

\begin{abstract}
 Scene depth estimation from paintings can streamline the process of 3D sculpture creation so that visually impaired people appreciate the paintings with tactile sense. However, measuring depth of oriental landscape painting images is extremely challenging due to its unique method of depicting depth and poor preservation.
To address the problem of scene depth estimation from oriental landscape painting images, we propose a novel framework that consists of two-step Image-to-Image translation method with CLIP-based image matching at the front end to predict the real scene image that best matches with the given oriental landscape painting image. Then, we employ a pre-trained SOTA depth estimation model for the generated real scene image.
In the first step, CycleGAN converts an oriental landscape painting image into a pseudo-real scene image.
We utilize CLIP to semantically match landscape photo images with an oriental landscape painting image for training CycleGAN in an unsupervised manner.
Then, the pseudo-real scene image and oriental landscape painting image are fed into DiffuseIT to predict a final real scene image in the second step.
Finally, we measure depth of the generated real scene image using a pre-trained depth estimation model such as MiDaS. 
Experimental results show that our approach performs well enough to predict real scene images corresponding to oriental landscape painting images. 
To the best of our knowledge, this is the first study to measure the depth of oriental landscape painting images. 
Our research potentially assists visually impaired people in experiencing paintings in diverse ways.
We will release our code and resulting dataset.
\end{abstract}

\vspace{-20px}

\begin{figure*}[t]
  \includegraphics[width=\linewidth,trim={1cm 3.5cm 0cm 3.5cm},clip]{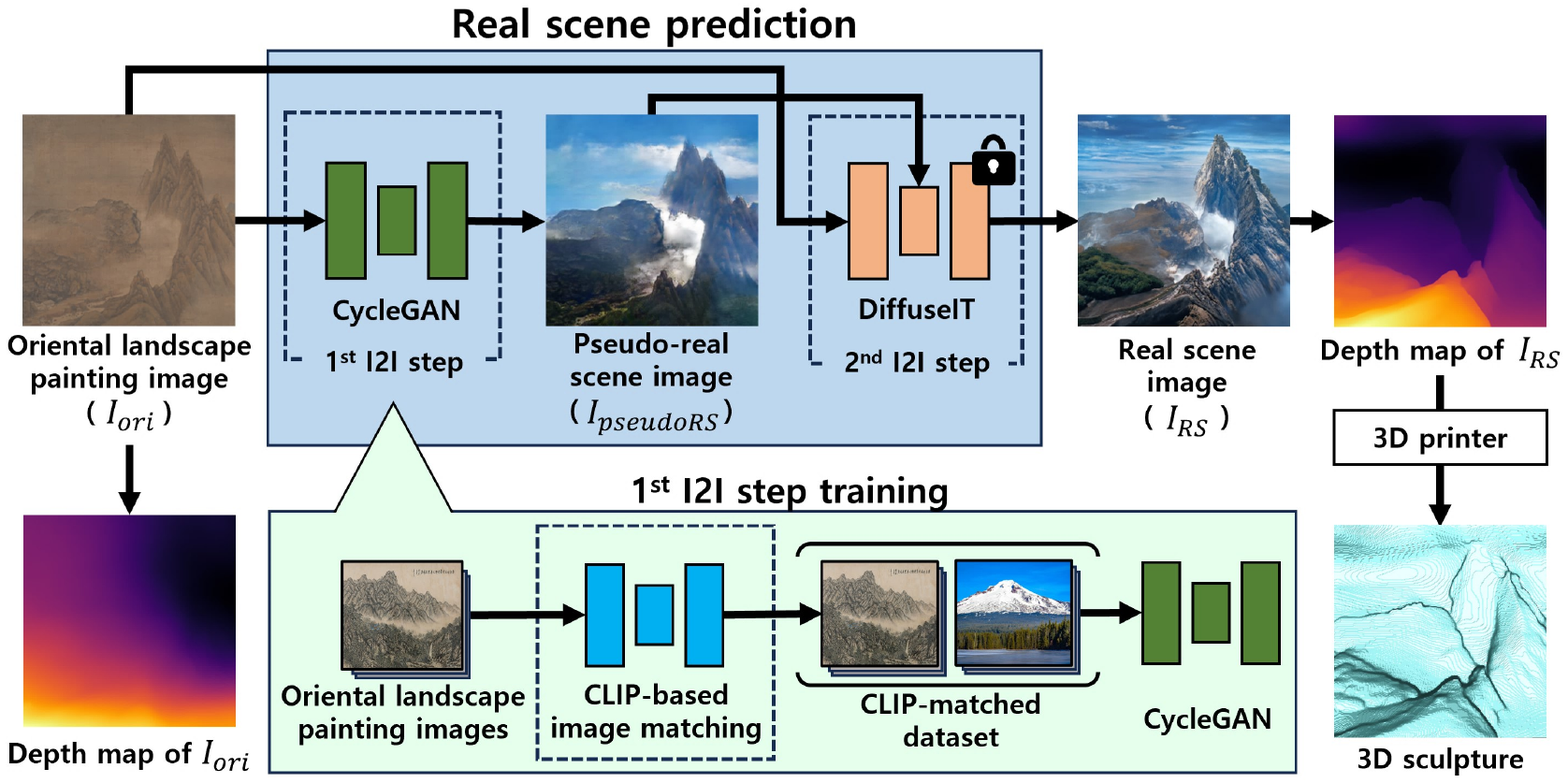} 
  \caption{An overview of our method. Direct application of pre-trained SOTA depth estimation models to the given oriental landscape painting image, $I_{ori}$, ends up with a meaningless depth map, `Depth map of $I_{ori}$'. 
We propose a novel framework that consists of CLIP-based image matching and I2I translation in two steps.
Through CLIP-based image matching, we initially obtain pairs of oriental landscape painting image and its corresponding landscape photo image.
For the I2I translation in two steps, we employ CycleGAN \cite{gi2i02} and DIffuseIT\cite{di2i01}. 
In the first step, CycleGAN \cite{gi2i02} that is trained on the CLIP-matched dataset translates an oriental landscape painting image into a pseudo-real scene image. 
To create a real scene image, the generated pseudo-real scene image and the given oriental landscape painting image are fed into DiffuseIT \cite{di2i01} in the second step.
The produced real scene image enables measuring depth using a pre-trained depth estimation model such as MiDaS \cite{MiDaS}. The measured depth map can be directly applied to craft 3D sculptures.}
\label{fig:goal}
\end{figure*}

\section{Introduction}
\label{sec:intro}
The advancement of media art technology enables visually impaired people to appreciate paintings with tactile sensation by crafting paintings as 3D sculptures\cite{mde09}. Motivated by this, we aim to find a way to produce depth maps particularly for traditional oriental landscape paintings in an attempt to automate the process of crafting paintings as 3D sculptures. Using the estimated depth maps, accurate 3D sculptures of the artwork are created in an efficient manner with the help of a 3D printer. 

\begin{figure*}[t]
  \includegraphics[width=\linewidth,trim={0cm 5.6cm 0cm 5.5cm},clip]{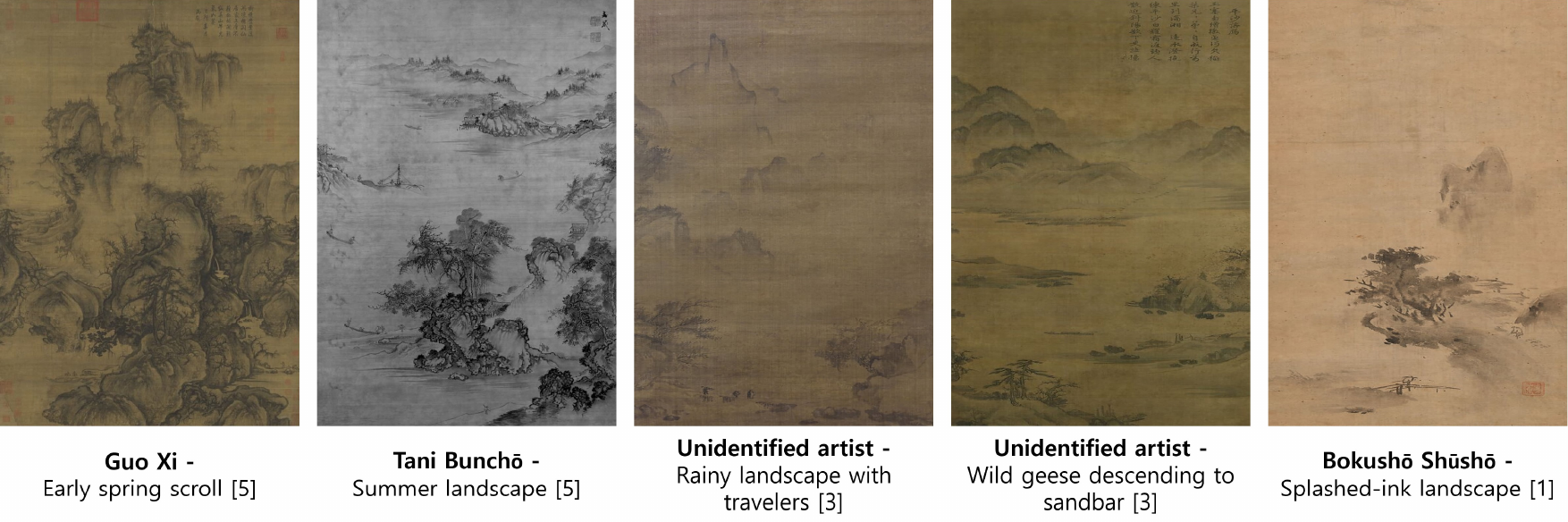} 
  \caption{Examples of oriental landscape paintings from museums \cite{mus01,mus02,mus04}. Oriental landscape paintings have been expressed in various styles due to their long history\cite{oriental09,oriental10}. An unique technique called `three-way method' is employed to create a sense of perspective\cite{oriental02,oriental04,oriental06,oriental07}.
  However, oriental landscape paintings often lack consistency in portraying objects due to the use of multiple types of 'three-way method' in a single painting. Moreover, oriental landscape paintings often contain empty spaces to depict depth of paintings\cite{oriental11,oriental05}. Additionally, oriental landscape paintings exhibit poor preservation conditions.}
\label{fig:oriental}
\end{figure*}

Depth estimation for oriental landscape painting images is a very challenging problem while pre-trained state-of-the-art (SOTA) depth estimation models can measure depth reasonably well in some western landscape painting images. 
Through the study of numerous oriental art papers \cite{oriental01,oriental02,oriental03,oriental04,oriental05,oriental06,oriental07,oriental08,oriental09,oriental10,oriental11,oriental12}, we have found why direct application of pre-trained SOTA depth estimation methods to oriental landscape painting images does not work. 
First, as illustrated in Fig. \ref{fig:oriental}, they use a unique method of depicting depth in oriental landscape paintings called `three-way method' that does not follow the perspective rule. This is the particular method of creating a sense of perspective by representing objects based on specific perspectives\cite{oriental02,oriental04,oriental06,oriental07}.
Also, painters often use lots of empty spaces to emphasize the semantic relationship between objects as a way to depict depth of oriental landscape paintings \cite{oriental11,oriental05}. 
Second, oriental landscape paintings have a long historical background, profoundly influenced and shaped by various nations including China, Japan, and Korea. This causes a multitude of stylistic variations\cite{oriental09,oriental10}.
Third, numerous oriental landscape paintings suffer from poor preservation, affected by blurry edges and weak contrast.

Thus, direct application of pre-trained SOTA depth estimation models to traditional oriental landscape painting images is likely to end up with meaningless depth maps as depicted in Fig. \ref{fig:goal}.
Because those depth estimation models are trained with real scene images which are in a huge domain difference from oriental landscape painting images.

Our strategy to achieve the best possible scene depth estimation is to first predict real scene images that best match given oriental landscape painting images, and then utilize a pre-trained SOTA depth estimation model to conduct depth estimation for the generated real scene images. 
The process of predicting a real scene image consists of two Image-to-Image (I2I) translation steps in an attempt to produce a high-quality real scene image corresponding to the given oriental landscape painting image for depth estimation. 
That is, in the first step, we employ CycleGAN\cite{gi2i02} to translate an oriental landscape painting image into a pseudo-real scene image that is input as a reference image for DiffuseIT\cite{di2i01}.
In the second step, the generated pseudo-real scene image and the oriental landscape painting image are fed into DiffuseIT\cite{di2i01} to create a real scene image in high-quality to estimate depth using a pre-trained SOTA depth estimation model.
The reason for utilizing the two I2I translation steps is that while diffusion-based I2I translation models show excellent performance in translating high-quality images, training two image domains with huge domain gap for diffusion-based I2I translation model does not work well.
Furthermore, if images for the two domains are structurally different, the output undergoes significant distortion during translation. 
Therefore, to utilize a pre-trained diffusion-based I2I translation model, for getting high quality images, it is necessary to produce pseudo-real scene images that are structurally similar to oriental landscape painting images.

More importantly, since actual location photo images matched with oriental landscape painting images are not available\cite{oriental01,oriental03}, we employ CLIP-based image matching to get pairs of oriental landscape painting images and corresponding landscape photo images for training CycleGAN\cite{gi2i02} to predict pseudo-real scene images for the given oriental landscape painting images.
For this, we semantically match oriental landscape painting images with landscape photo images by using a pre-defined dictionary. 
To build the pre-defined dictionary, we have carefully selected objects that frequently appear in oriental landscape paintings by referring to museum collections\cite{mus01,mus02,mus03,mus04,mus05} and oriental art papers\cite{oriental08,oriental12}.

Our method can be summarized as follows. 
We first build a CLIP-matched dataset by matching oriental landscape painting images from the Chinese landscape dataset \cite{OLP} with semantically similar landscape photo images from the LHQ dataset \cite{LHQ}, based on the pre-defined dictionary using CLIP \cite{CLIP01}. When creating the CLIP-matched dataset, a 1-to-$K$ matching manner is employed that matches a given oriental landscape painting image with the $K$ most similar landscape photo images, instead of a 1-to-1 matching manner. This reduces the impact on the performance when CLIP\cite{CLIP01} incorrectly matches oriental landscape painting images with landscape photo images.
CycleGAN \cite{gi2i02}, trained on the CLIP-matched dataset, converts oriental landscape painting images into pseudo-real scene images. Then, the oriental landscape painting images and the pseudo-real scene images are fed into the pre-trained DiffuseIT \cite{di2i01} to create real scene images that are structurally similar to the oriental landscape painting images.
Finally, depth of real scene images is measured using pre-trained MiDaS\cite{MiDaS}. 
MiDaS\cite{MiDaS} is one of the SOTA depth estimation models, which is trained on variety of datasets, showing an outstanding generalization performance that leads to produce plausible depth maps of paintings with high probability.
The resulting depth maps of real scene images can be directly used in crafting 3D sculptures, which can assist visually impaired people in appreciating paintings with their touch. 
The experimental results show that our method outperforms previous I2I translation methods in predicting real scene images corresponding to oriental landscape painting images.
Accordingly, the predicted real scene images enable MiDaS\cite{MiDaS} to measure depth in a satisfactory manner. 
To the best of our knowledge, this is the first study to measure the depth of oriental landscape painting images.

Our contributions are as follows.
\begin{itemize}
\item We introduce a novel framework that converts oriental landscape painting images into real scene images using I2I translation models in two steps to produce best possible real scene images of high-quality for depth estimation.
\item We propose a CLIP-based image matching method to initially obtain pairs of oriental landscape painting image and its corresponding landscape photo image. For this, we have built a pre-defined dictionary by selecting objects that frequently appear in oriental landscape paintings by referring to museum collections and oriental art papers. This strategy gives a new perspective to an otherwise very difficult problem with no image pairs.
\end{itemize}

\section{Related work}

\subsection{Depth estimation in painting}
Various depth estimation methods were proposed and showed satisfactory performances in data domains such as synthetic data\cite{mde03,mde04}, multi-view cameras\cite{mde02}, and LiDAR\cite{mde01}. Especially, MiDaS\cite{MiDaS} showed excellent generalization performance, allowing their network to estimate depth in images completely unrelated to the training dataset such as paintings. Ranftl \textit{et al}.\cite{DPT} changed the architecture of MiDaS\cite{MiDaS} to transformer-based encoder-to-decoder architecture to achieve better performance. Birkl \etal.\cite{zoo} presented variety of new models of MiDaS\cite{MiDaS} based on different pre-trained vision transformers. Bhat \textit{et al}.\cite{mde07} and Miangoleh \textit{et al}.\cite{mde08} produced more detailed depth maps based on the approach of MiDaS\cite{MiDaS}.

On the other hand, there have been several attempts to leverage deep learning models to measure the depth of painting images. Jiang \textit{et al}.\cite{mde10} utilized a depth estimation model\cite{mde11} to gauge the depth in western painting images. Lin \textit{et al}. \cite{mde12} introduced a user-interactive framework to generate depth maps of painting images and photo images manually. Bhattacharjee \textit{et al}.\cite{mde05} proposed a novel framework that estimates the depth of comics by transforming them into real scenes. 
Pauls \textit{et al}. used depth maps of painting images to enhance the experience of museum visits\cite{mde09}.
Ehret \textit{et al}.\cite{mde06} reported that although DPT \cite{DPT} and MiDaS \cite{MiDaS} were not explicitly trained for measuring the depth of painting images, they produced reasonable depth maps based on the high generalization performance.

Previous depth estimation works in painting images only attempted to measure the depth of western landscape painting images that are somewhat similar to landscape photo images. We have observed SOTA depth estimation model such as MiDaS \cite{MiDaS} cannot generate satisfactory depth maps for oriental landscape painting images due to their characteristics mentioned earlier.
We take the strategy that first converts oriental landscape painting images into real scene images and then measures depth of real scene images using MiDaS\cite{MiDaS}.

\subsection{Image-to-Image translation from painting to photo}
I2I translation is a popularly used solution to transformation of an image in domain A into an image in domain B. Tomei \textit{et al}.\cite{gi2i06} proposed a method to convert painting images into photo images. This method employed a segmentation model\cite{mde13} to divide painting images and photo images into patches and then match them, showing an outstanding translation performance. Despite its good performance, it cannot be applied to convert oriental landscape painting images to real scene images due to the huge domain gap between them.

Pix2Pix \cite{gi2i01} was the first to apply conditional GAN to the I2I translation task and showed excellent performances. On the other hand, CycleGAN\cite{gi2i02} was proposed as a way for unsupervised learning to the I2I translation. 
Park \textit{et al}.\cite{gi2i05} presented a method, dubbed CUT, that maximizes mutual information based on contrastive learning.
Liu \textit{et al}. \cite{gi2i04} showed a shared-latent space assumptions and proposed a framework, called UNIT.
Lee \textit{et al}.\cite{gi2i11} introduced a disentangled representation framework with cross-cycle consistency loss.
Recently, Chen \textit{et al}.\cite{gi2i09} conducted I2I translation based on VQGAN\cite{gi2i10} which gives an outstanding performance in image generation.

Although GAN-based models performed well in I2I translation tasks, Ho \textit{et al}.\cite{DDPM} successfully applied diffusion process to neural network, yielding many diffusion-based I2I models that translate images with high-quality.
Kwon \textit{et al}.\cite{di2i01} proposed diffusion-based I2I and Text-to-Image (T2I) translation methods using disentangled style and content representation. 
Saharia \textit{et al}. \cite{di2i02} implemented diffusion models for various tasks, including I2I translation, without task-specific hyperparameter tuning. 
Su \textit{et al}.\cite{di2i03} proposed diffusion-based I2I translation in latent encoding and defined it via ordinary differential equations. 
Li \textit{et al}.\cite{di2i04} successfully applied stochastic brownian bridge process to diffusion for translation between two domains directly through the bidirectional diffusion process.

GAN-based I2I translation models cannot realistically transform oriental landscape painting images into landscape photo images as diffusion-based models do. 
However, diffusion-based models are not directly applicable to our problem because they drastically modify structural information of image during translation when input images are structurally different.
Therefore, to overcome these limitations, we employ GAN-based model in the first I2I step and diffusion-based model in the second I2I step.

\subsection{CLIP-based image matching}
\label{clip-based mathing}
Radford \textit{et al}.\cite{CLIP01} proposed a contrastive learning-based Image-to-Text (I2T) matching model which allows zero-shot prediction called CLIP. When CLIP\cite{CLIP01} gets an input image and a set of text, it finds the text that best matches the image in multi-modal embedding space. Diverse CLIP-based methods are proposed, for image retrieval\cite{CLIP03,CLIP04}, style transfer\cite{CLIP05,CLIP06}, image generation\cite{CLIP06,CLIP07,CLIP08}, classification\cite{CLIP10,CLIP11}, anomaly detection\cite{CLIP12}, and image aesthetic assessment\cite{CLIP15,CLIP16}. Vinker \textit{et al}.\cite{CLIP13} introduced a technique for converting images into sketches by employing diverse types and multiple levels of abstraction. Hessel \textit{et al}.\cite{CLIP14} proposed a novel metric for robust automatic evaluation of image captioning. Especially, Materzyńska \textit{et al}.\cite{CLIP02} found that CLIP\cite{CLIP01} had a strong ability of matching words with natural images of scenes.

To generate more plausible real scene images, we build a CLIP-matched dataset consisting of pairs of oriental landscape painting images and landscape photo images since there are no actual location photo images that correspond to oriental landscape painting images. 
We opt to match semantically similar images to a given oriental landscape painting images.
For this, we utilized CLIP\cite{CLIP01} for building the CLIP-matched dataset because CLIP\cite{CLIP01} may allow matching tasks in unseen image domains such as oriental landscape painting images to landscape photo images.

\begin{figure*}[ht]
  \includegraphics[width=\linewidth,trim={0cm 3cm 0cm 2cm},clip]{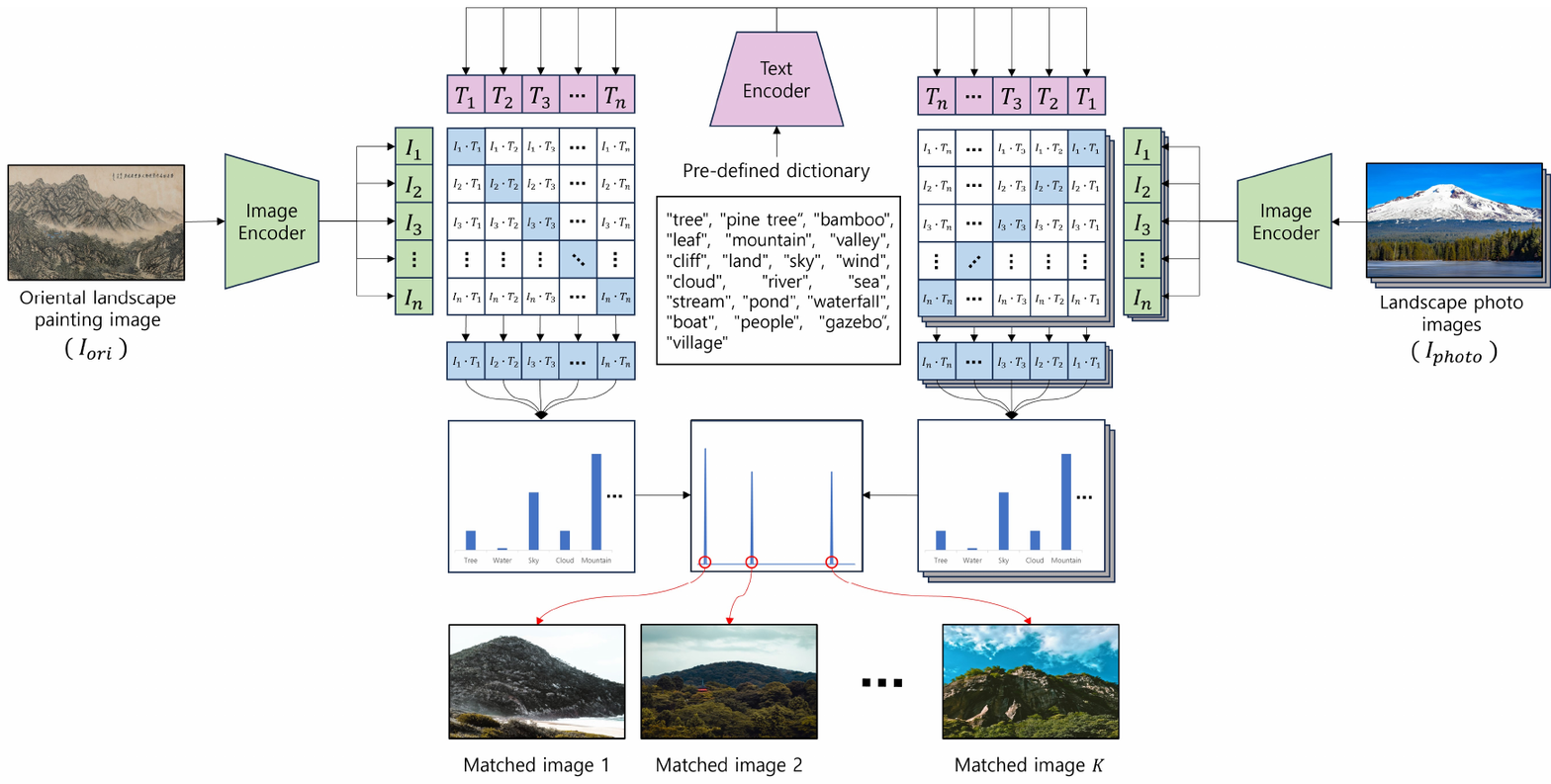}  
  \caption{An image matching method using CLIP\cite{CLIP01}. We employ CLIP \cite{CLIP01} to semantically match oriental landscape painting images with landscape photo images. To build a pre-defined dictionary for CLIP \cite{CLIP01}, we carefully selected frequently appearing objects in oriental landscape paintings by referencing papers \cite{oriental08,oriental12} and collections from museums \cite{mus01,mus02,mus03,mus04,mus05}. The pre-defined dictionary is fed into the text encoder of CLIP\cite{CLIP01}, while oriental landscape painting images are input into the image encoder to measure similarity. Simultaneously, all landscape photo images in the LHQ dataset \cite{LHQ} are fed into another image encoder for measuring similarity. By comparing the similarity, we match an oriental landscape painting image with the top-$K$ landscape photo images most similar to them.
  The top-$K$ most similar landscape photo images matched to the given oriental landscape painting images are used to create the CLIP-matched dataset for training CycleGAN \cite{gi2i02} to generate more plausible pseudo-real scene images.}
\label{fig:clipmatching}
\end{figure*}

\vspace{5px}

\section{Method}
As depicted in Fig. \ref{fig:goal}, our method exploits CLIP-based image matching and two-step I2I translation to predict 
a high-quality real scene image corresponding to a given oriental landscape painting image. For the I2I translation in two steps, CycleGAN\cite{gi2i02} and DiffuseIT\cite{di2i01} are sequentially employed.
In the first step, we utilized CycleGAN\cite{gi2i02} to get a pseudo-real scene images that is structurally similar to a given oriental landscape painting image.
In the second step, DiffuseIT\cite{di2i01} predicts a real scene image of high-quality using the generated pseudo-real scene image and the given oriental landscape painting image. The predicted real scene image is utilized for depth estimation by pre-trained MiDaS\cite{MiDaS}.

\vspace{5px}

\subsection{CLIP-matched dataset}
Since oriental landscape painting images, $I_{ori}$, and landscape photo images, $I_{photo}$, are in different domains, direct I2I matching in the embedding space does not perform well. 
Furthermore, actual location photo images that matches with oriental landscape painting images are not available. 
Thus, we utilize CLIP\cite{CLIP01} to semantically match oriental landscape painting images with landscape photo images for building a CLIP-matched dataset.

To make the CLIP-matched dataset, we built a pre-defined dictionary.
As the pre-defined dictionary is important for matching, we have selected frequently appearing objects in $I_{ori}$ by referencing papers \cite{oriental08,oriental12} and museum collections \cite{mus01,mus02,mus03,mus04,mus05}. 
Park \textit{et al}. \cite{oriental12} reported that the analysis of 629 oriental landscape paintings showed that 98.4\% of them depicted mountains, rivers, and the sea, underlining that many oriental landscape paintings are related to nature, while those depicting villages are scarce. 
Therefore, when constructing a pre-defined dictionary, it is reasonable to include more words related to nature.
According to Kim \textit{et al}. \cite{oriental08}, when expressing water in oriental landscape paintings, it is subdivided into details such as waterfall, stream, and pond based on the terrain. 

Hence, when creating the pre-defined dictionary, we not only included keywords related to nature, but also categorized those keywords.
To semantically match $I_{ori}$ with $I_{photo}$ using CLIP \cite{CLIP01}, we build the pre-defined dictionary as shown in Fig. \ref{fig:clipmatching}.

Each set of images has $n$ and $m$ samples of $I_{ori}$ or $I_{photo}$ as in ~\eqref{eq:set_of_image}. 
\begin{equation}
    \begin{aligned}
        I_{ori} = \{ i_{ori}^{1}, \cdots i_{ori}^{m} \}, \ 
        I_{photo} = \{ i_{photo}^{1}, \cdots i_{photo}^{n} \}, \ 
    \end{aligned}
    \label{eq:set_of_image}
\end{equation}

To avoid limitation that CLIP-based image matching method incorrectly matched $I_{ori}$ with $I_{photo}$, we construct a CLIP-matched dataset using a 1-to-$K$ matching method instead of a 1-to-1 matching manner. Thus, for each $I_{ori}$, we search top-$K$ best matching results in a set of $I_{photo}$.
The result of CLIP-based matching will be $I'_{photo}$, a subset of $I_{photo}$, as expressed in~\eqref{eq:set_of_real}.
\begin{equation}
    \begin{aligned}
        S_{photo}^{K}(i_{ori}^{j})_{j={1, \cdots m}} = I'_{photo}, \ I'_{photo} \subset I_{photo}
    \end{aligned}
    \label{eq:set_of_real}
\end{equation}
In summary, since there are no actual location photo images corresponding to oriental landscape painting images, we employed CLIP\cite{CLIP01} to match landscape photo images that are semantically similar to oriental landscape painting images. The CLIP-matched dataset that consists of matched images is used to train CycleGAN \cite{gi2i02} in an unsupervised manner. The performance evaluation for different values of $K$ in 1-to-$K$ matching is presented in the Section \ref{k in clip-based}.
\vspace{5px}

\subsection{Two-step real scene image prediction with I2I translation}
As in CycleGAN\cite{gi2i02}, we employ two auto-encoders, $G_{photo\rightarrow ori}$ and $G_{ori \rightarrow photo}$, for mapping. $G_{photo\rightarrow ori}$ transforms $I_{photo}$ into $I_{ori}$, while $G_{ori \rightarrow photo}$ converts $I_{ori}$ into $I_{photo}$. Additionally, we use two discriminators, $D_{photo}$, and $D_{ori}$. $D_{photo}$ learns to differentiate $G_{photo\rightarrow ori}(I_{ori})$ and $I_{photo}$ as fake and real, while $D_{ori}$ distinguishes between $G_{ori \rightarrow photo}(I_{photo})$ and $I_{ori}$, aiming to discern fake and real images.

The utilized adversarial and consistency loss can be summarized as follows: The adversarial loss ensures that the auto-encoder $G_{photo\rightarrow ori}$ generates images as similar as possible to $I_{ori}$ and simultaneously trains the discriminator $D_{ori}$ to classify $I_{ori}$ as real and $G_{photo\rightarrow ori}(I_{ori})$ as fake. This process facilitates the transformation of $I_{photo}$ into $I_{ori}$ as closely as possible.

\begin{equation}
    \begin{aligned}
        L_{adv}&(G_{photo\rightarrow ori},D_{ori},I_{ori},I_{photo})\\
        & =\mathbb{E}_{I_{ori}\sim P_{data}(I_{ori})}\left [log(D_{ori}(I_{ori})) \right ] \, \\
        & \quad  + \mathbb{E}_{I_{photo}\sim P_{data}(I_{photo})}\left [log(1-D_{ori}(G_{photo\rightarrow ori}(I_{photo}))) \right ]
    \end{aligned}
    \label{advloss}
\end{equation}
\vspace{5px}
Adversarial loss is also employed during the transformation from $I_{ori}$ to $I_{photo}$, is expressed as:

\begin{equation}
    \begin{aligned}
        L_{adv}&(G_{ori \rightarrow photo},D_{photo},I_{ori},I_{photo})\\
        & =\mathbb{E}_{I_{photo}\sim P_{data}(I_{photo})}\left [log(D_{photo}(I_{photo}))\right ] \, \\ 
        & \quad +\mathbb{E}_{I_{ori}\sim P_{data}(I_{ori})}\left [log(1-D_{photo}(G_{ori \rightarrow photo}(I_{ori}))) \right ]
    \end{aligned}
    \label{advloss2}
\end{equation}
\vspace{5px}
The cycle consistency loss is utilized under the premise that when $I_{photo}$ is transformed using $G_{photo\rightarrow ori}$ and the resulting $G_{photo\rightarrow ori}(I_{photo})$ is further transformed using $G_{ori \rightarrow photo}$, the output $G_{ori \rightarrow photo}(G_{photo\rightarrow ori}(I_{photo}))$ should resemble $I_{photo}$ when case of $I_{ori}$ should be same. The equation can be represented as:

\begin{equation}
    \begin{aligned}
L_{cyc}&(G_{photo\rightarrow ori},G_{ori \rightarrow photo})\\
& =\mathbb{E}_{I_{photo}\sim P_{data}(I_{photo})} \left [ \left\|G_{ori \rightarrow photo}(G_{photo\rightarrow ori}(I_{photo}))-I_{photo} \right\|_{1} \right ] \,\\
& \quad + \mathbb{E}_{I_{ori}\sim P_{data}(I_{ori})}\left [ \left\|G_{photo\rightarrow ori}(G_{ori \rightarrow photo}(I_{ori}))-I_{ori} \right\|_{1} \right ]
    \end{aligned}
    \label{cycloss}
\end{equation}
\vspace{5px}
The final objective loss term can be expressed as follows, with each term adjusted using $\lambda$s.

\begin{equation}
    \begin{aligned}
L_{total}&(G_{photo\rightarrow ori},G_{ori\rightarrow photo},D_{ori},D_{photo})\\
& = L_{adv}(G_{photo\rightarrow ori},D_{ori},I_{ori},I_{photo}) \, \\
& \quad \; + \lambda_{adv} L_{adv}(G_{ori\rightarrow photo},D_{photo},I_{ori},I_{photo}) \, \\
& \quad \; + \lambda_{cyc} L_{cyc}(G_{photo\rightarrow ori},G_{ori\rightarrow photo})
 \end{aligned}
    \label{fullobj}
\end{equation}

In summary, in the first step, we use CycleGAN\cite{gi2i02} to transform $I_{ori}$ into $I_{photo}$. We denote the translated $I_{photo}$ as $I_{pseudoRS}$, \ie, $\Phi_{CycleGAN}(I_{ori}) = I_{pseudoRS}$.
The generated $I_{pseudoRS}$ and $I_{ori}$ are fed into a pre-trained DiffuseIT\cite{di2i01} to predict a real scene image, $I_{RS}$, \ie, $\Phi_{DiffuseIT}(I_{ori},I_{pseudoRS}) = I_{RS}$.
The generated $I_{RS}$ enables depth prediction via a pre-trained depth estimation model such as MiDaS \cite{MiDaS}.

\begin{figure*}[t!]
  \includegraphics[width=\linewidth,trim={0.2cm 0cm 0.2cm 0cm},clip]{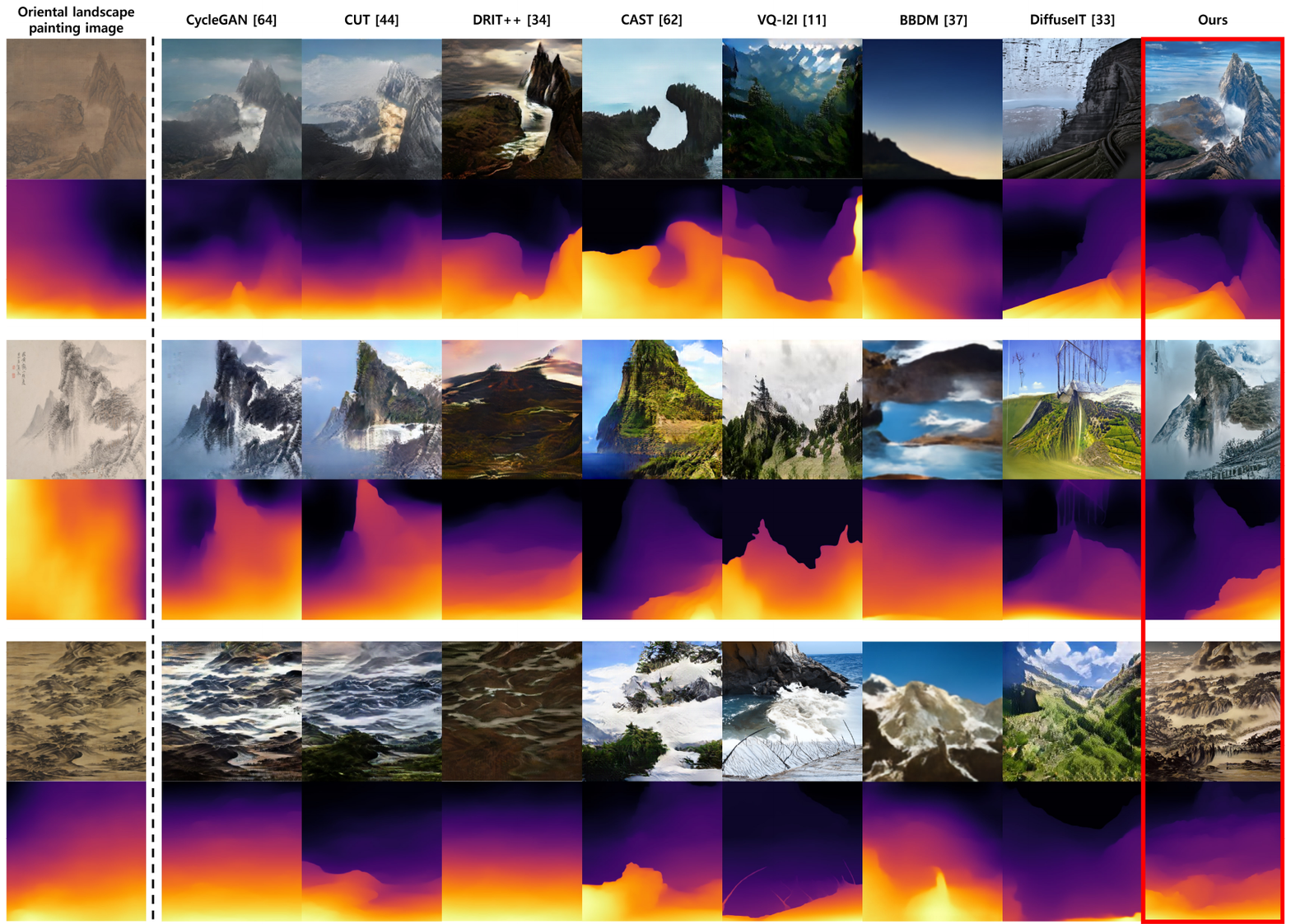}
  \caption{For qualitative evaluation, we compared our method with other I2I translation models. Initially, MiDaS \cite{MiDaS}, a pre-trained depth estimation model, failed to accurately estimate the depth of oriental landscape painting images despite its strong generalization ability. While GAN-based I2I translation methods \cite{gi2i02,gi2i05,gi2i11} preserved the structural information of these painting images, they struggled to achieve realistic translations. CAST \cite{gi2i08} and VQ-I2I \cite{gi2i09} not only failed to produce realistic translations but also lost structural fidelity. 
Diffusion-based I2I models such as DiffuseIT \cite{di2i01} and BBDM \cite{di2i04} managed to realistically transform oriental landscape painting images into real scene images, albeit with huge distortion in structural information.  
In contrast, our method preserve structural information like CycleGAN \cite{gi2i02}, CUT\cite{gi2i05} or DRIT++ \cite{gi2i11} while achieving realistic translations akin to BBDM \cite{di2i04} or DiffuseIT\cite{di2i01}. Our approach successfully enables depth measurement using a pre-trained MiDaS\cite{MiDaS} for the given oriental landscape painting images.}
\label{fig:qualitativeresult}
\end{figure*}

\begin{figure*}[t]
  \includegraphics[width=\linewidth,trim={0cm 5cm 0cm 5cm},clip]{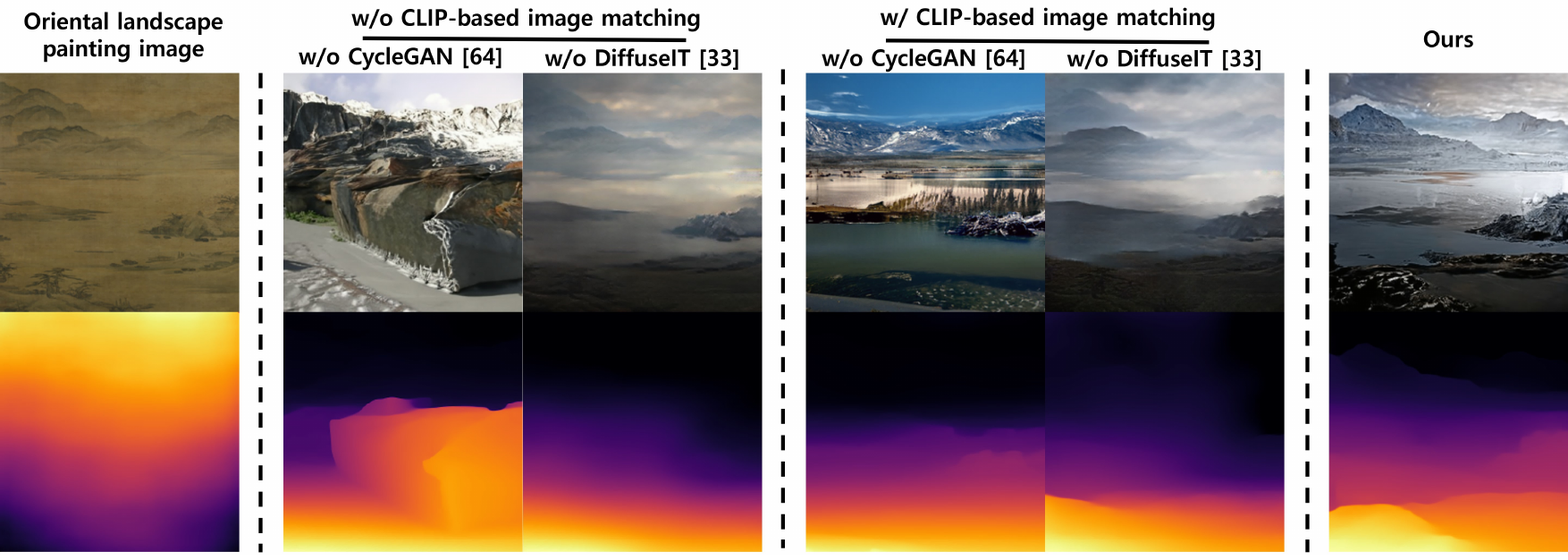} 
  \caption{The w/o CycleGAN\cite{gi2i02} cases show significant structural distortion during translation. When DiffuseIT\cite{di2i01} is missing, predicted images maintain structure of oriental landscape painting images but it fails to realistically convert to real scene images.
The w/ CLIP-based image matching cases demonstrate that CLIP-based image matching make CycleGAN\cite{gi2i02} for producing high-quality pseudo-real scene image.
Our method shows that employing CLIP-based image matching, CycleGAN\cite{gi2i02}, and DiffuseIT\cite{di2i01} generate more plausible real scene images. Best viewed in color.}
\label{fig:ablation}
\end{figure*}

\section{Experiments}
\vspace{-8px}
To evaluate the performance of our method, we provide experimental settings, qualitative evaluation, ablation study, optimal $K$ in CLIP-based image matching, and user study. 

In the Section \ref{qualitative}, we have compared our method with previous I2I translation models\cite{gi2i02,gi2i05,gi2i08,gi2i09,gi2i11,di2i01,di2i04} for qualitative evaluation.
In addition, the ablation study is also provided to demonstrate the importance of each part of the proposed framework in the Section \ref{qualitative}. 

For the experimental settings, as the first step of I2I, CycleGAN\cite{gi2i02} has been trained in an unsupervised manner using the CLIP-matched dataset that paired oriental landscape painting images from Chinese landscape dataset \cite{OLP} with the corresponding landscape photo images from LHQ dataset \cite{LHQ}. For DiffuseIT\cite{di2i01} in the second I2I step, ImageNet $256\times{}256$ pre-trained model has been utilized. Other I2I translation models\cite{gi2i02,gi2i05,gi2i08,gi2i09,gi2i11,di2i01,di2i04} have been trained in an unsupervised manner, using the Chinese landscape dataset \cite{OLP} as oriental landscape painting images and random sampled of LHQ dataset \cite{LHQ} as landscape photo images. Their official source codes and recommended hyperparameters have been used for training. For generating depth maps, BEiT$_{512}$-L pre-trained MiDaS v3.1\cite{MiDaS} has been employed to estimate depth. 

In order to analyze the impact of $K$ on CLIP-based image matching and find the optimal $K$ value, we have experimented with different $K$ values, \ie, $K$ = 1, 3, 5, and 10.

Since there is no metric for quantitative evaluation, we conducted a user study. 
We asked forty anonymous participants five questions, each question contains $Q^s$ and $Q^q$. 
$Q^s$ aims to to assess how well the real scene image preserves the structure of a given oriental landscape painting image during translation. $Q^q$ is for rating how well a produced real scene image is realistically translated. Detailed information is provided in following section.

\begin{figure*}[t]
  \includegraphics[width=\linewidth,trim={0cm 6.5cm 0cm 6.5cm},clip]{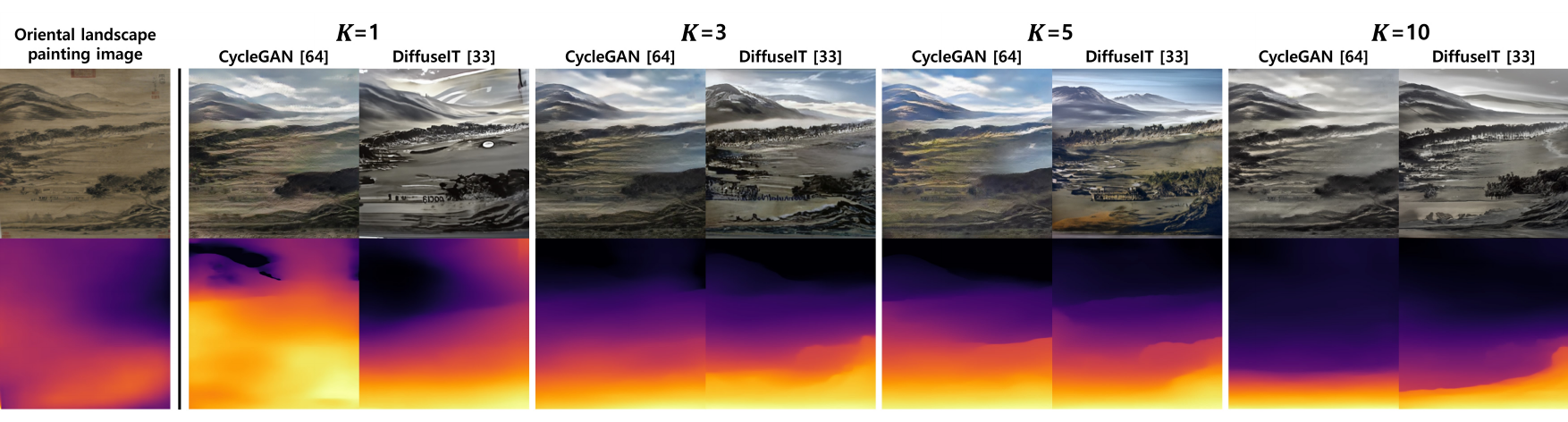} 
  \caption{When the $K$ is either 1 or 10, CycleGAN \cite{gi2i02} has not been trained effectively, a pseudo-real scene image of low-quality is generated. Consequently, DiffuseIT \cite{di2i01} failed to predict a plausible real scene image in the second step, and its outcome is also linked to depth measurements. When the $K$ is set to either 3 or 5, in the first step translation, a pseudo-real scene image is generated more realistically. The second step I2I translation also predicts the best real scene image that corresponds to a given oriental landscape painting image. This result demonstrates that suitable $K$ in CLIP-based image matching leads to produce more plausible real scene image. Furthermore, the quality of the generated real scene image affects the depth estimation results.}
  \label{fig:k-factor}
\end{figure*}

\subsection{Qualitative evaluation}
\label{qualitative}
To evaluate qualitative results, we have compared our method with other I2I translation models.
For our methods, CycleGAN\cite{gi2i02} in the first step has been trained on the CLIP-matched dataset. ImageNet $256\times{}256$ pre-trained model has been employed for DIffuseIT\cite{di2i01} in the second step.
Other I2I translation models have been trained on pairs with oriental landscape painting images from Chinese landscape painting dataset\cite{OLP} and randomly sampled landscape photo images from LHQ dataset\cite{LHQ}.

As illustrated in the first column of Fig. \ref{fig:qualitativeresult}, direct application of pre-trained depth estimation model, MiDaS \cite{MiDaS}, fails to estimate depth of oriental landscape painting images although MiDaS\cite{MiDaS} has excellent generalization ability.
As depicted in the second, third, and fourth columns of Fig. \ref{fig:qualitativeresult}, the GAN-based I2I translation methods\cite{gi2i02,gi2i05,gi2i11} well preserve the structural information of the oriental landscape painting images, but cannot convert them realistically.
Surprisingly, the GAN-based I2I translation models such as CAST\cite{gi2i08} and VQ-I2I\cite{gi2i09} not only fail to realistically translate oriental landscape painting images to real scene images, but also distort structural information as shown in the fifth and sixth columns of Fig. \ref{fig:qualitativeresult}.
In the seventh and eighth columns of Fig. \ref{fig:qualitativeresult}, DiffuseIT \cite{di2i01} and BBDM \cite{di2i04}, diffusion-based I2I models realistically transform oriental landscape painting images into real scene images. However, predicted real scene images drastically change structural information during translation. 
This means that even if depth can be estimated using pre-trained MiDaS\cite{MiDaS}, obtained depth map does not represent the scene in the given oriental landscape painting images.
Our method predicts real scene images corresponding to oriental landscape painting images, preserving structural information such as CycleGAN \cite{gi2i02}, CUT\cite{gi2i05} or DRIT++ \cite{gi2i11} and realistically translating into real scene images, like BBDM \cite{di2i04} or DiffuseIT \cite{di2i01}.

As depicted in Fig. \ref{fig:ablation}, regardless of the use of CLIP-based image matching, absence of CycleGAN\cite{gi2i02} leads to significant structural deformation during translation. Depth of structurally modified real scene images is meaningless even if depth can be measured using a pre-trained depth estimation model\cite{MiDaS}.
In cases that DiffuseIT\cite{di2i01} is missing, although the predicted real scene images maintain structural similarity, it fails to realistically convert oriental landscape painting images into real scene images that lead to unsatisfactory depth map outcomes. 
Without CLIP-based image matching, it creates low-quality pseudo-real scene images which affect the final result.
To achieve plausible real scene images corresponding to given oriental landscape painting images, CLIP-based matching, CycleGAN\cite{gi2i02} and DiffuseIT\cite{di2i01} should be employed.

\subsection{Optimal $K$ in CLIP-based matching}
\label{k in clip-based}
As illustrated in Fig. \ref{fig:k-factor}, we have trained CycleGAN \cite{gi2i02} using the CLIP-matched dataset.
When building this dataset, a 1-to-$K$ manner is employed. To prove effectiveness of the 1-to-$K$ manner, we have compared real scene images with $K$ as 1, 3, 5, and 10. 
Setting $K$ to either 1 or 10 results in either too few or too many images treated as correct data, significantly influencing CycleGAN \cite{gi2i02} and ultimately yielding poor translated outcomes. When $K$ is set to either 3 or 5, CycleGAN \cite{gi2i02} produces the most plausible pseudo-real scene images, which also affect the final results through DiffuseIT \cite{di2i01}.

\subsection{User study}
\label{userstudy}
To evaluate structural preservation and translation quality of predicted real scene image, we conducted a user study. We asked forty anonymous participants five question sets that are subdivided to $Q^s$ and $Q^q$.

\begin{table}[h]
\centering
\caption{User study result. For $Q^s$, more than 90\% of the respondents chose the oriental landscape painting images that correspond to the real scene image for all the questions, with an average of 93.6\%. This shows that our method effectively preserves structural information during transformation.
Additionally, for the questions regarding the quality of the predicted real scene images, $Q^q$, all the questions received scores of higher than 4.0, with an average score of 4.22. This indicates that the predicted real scene images are realistically translated.}
\label{tab:userstudy}
\begin{tabular}{cc|c|c|c|c|c|c}
\hline
\multicolumn{2}{l|}{Question No.} & 1    & 2     & 3    & 4    & 5    & avg    \\ \hline \hline
$Q^s$     & score (\%)           & 90\% & 100\% & 93\% & 95\% & 90\% & 93.6\% \\
$Q^q$     & \, \, score (1$\sim$5)     & 4.3  & 4.3   & 4.1  & 4.2  & 4.2  & 4.22   \\ \hline
\end{tabular}
\end{table}

First, for $Q^s$, we provided one real scene image and five oriental landscape painting images that include the one oriental landscape painting image corresponding to the given real scene image.
We asked participants to select a single oriental landscape painting image that is most structurally similar to the real scene image.
This question aims to assess how well the real scene image preserves the structure of the given oriental landscape painting image during translation.

After the first question was done, we asked another question, $Q^q$. We gave participants a pair of an oriental landscape painting image and its real scene image and asked them to rate how well the produced real scene image is realistically translated, scaling from min 1 to max 5.

As summarized in Table \ref{tab:userstudy}, regarding the assessment of structural preservation in questions, $Q^s$, an average of 93.6\% of the participants chose oriental landscape painting images that corresponded to predicted real scene images. 
For the questions that evaluate to quality of the predicted real scene images, $Q^q$, an average score of 4.22 point was obtained. 
These results serve as evidence demonstrating the capability of our method to not only preserve the structural similarity of oriental landscape paintings during prediction, but also realistically translate oriental landscape painting image to real scene image.
\vspace{-2px}

\section{Conclusion}
In this paper, we propose a novel framework that features CLIP-based image matching and two-step I2I translation to predict real scene images corresponding to oriental landscape painting images. We utilize a CLIP-based image matching method with the pre-defined dictionary that includes objects commonly appearing in oriental landscape paintings.
In the first I2I step, oriental landscape painting images are translated to pseudo-real scene images using CycleGAN\cite{gi2i02}. 
The generated pseudo-real scene images and the given oriental landscape painting images are input into DiffuseIT\cite{di2i01} for predicting final real scene images in the second I2I step.
This strategy enables oriental landscape painting images to be translated into real scene images that are structurally similar yet realistic.

Experimental results prove to be highly effective when measuring the depth of oriental landscape painting images through I2I translations. 
The measured depth of the predicted real scene images can be directly applied when generating 3D sculptures that aid visually impaired people in experiencing paintings through touch. 
We hope that our work paves the way to help visually impaired people to appreciate diverse paintings.

There are tasks that need to be addressed in future research. oriental landscape paintings suffer from inherent preservation issues such as splitting, blurring, and deformation which significantly impact the outcomes of I2I translation. In other words, a model that restores oriental landscape paintings to their original state is required. Furthermore, oriental landscape paintings often contain numerous text and stamps that impact to depth estimation results. Removing these text, written with similar strokes and colors as the artwork, poses difficulties. To address this problems, substantial amounts of data need to be collected and shared. However, many globally recognized museums frequently restrict the public release and manipulation of their data.

\section{Acknowledgement}
This study was supported in part by the National Research Foundation of Korea (NRF) under Grant 2020R1F1A1048438 and the High-performance Computing (HPC) Support project funded by the Ministry of Science and ICT and National IT Industry Promotion Agency (NIPA). Also, we sincerely thank the anonymous participants who participated in the user study.


%
%
\bibliographystyle{splncs04}
\bibliography{main.bib}
\end{document}